\documentclass{article}

\usepackage[preprint]{neurips_2025}

\usepackage[pagebackref,breaklinks,colorlinks,citecolor=blue]{hyperref}

\usepackage[utf8]{inputenc} %
\usepackage[T1]{fontenc}    %
\usepackage{hyperref}       %
\usepackage{url}            %
\usepackage{booktabs}       %
\usepackage{amsfonts}       %
\usepackage{graphicx}       %
\usepackage{nicefrac}       %
\usepackage{microtype}      %
\usepackage{xcolor, eucal}         %
\usepackage{multirow}
\usepackage{amsmath}
\usepackage{caption}
\usepackage{siunitx}
\usepackage{enumitem}
\usepackage[most]{tcolorbox}
\tcbuselibrary{skins, breakable}
\usepackage{changepage}
\usepackage{subcaption}

\newcommand{\ours}{{\sc Om\-ni-\-R1}}

\title{Omni-R1: Reinforcement Learning for Omnimodal Reasoning via Two-System Collaboration}

\def\hsp{{\hspace{2mm}}}

\author{%
  Hao Zhong\thanks{Equal contribution.}, \hsp
  Muzhi Zhu\footnotemark[1], \hsp
  Zongze Du\footnotemark[1], \hsp
  Zheng Huang, \hsp
  Canyu Zhao, \\
  \bf Mingyu Liu, \hsp
  Wen Wang, \hsp
  Hao Chen, \hsp
  Chunhua Shen\thanks{C. Shen is the corresponding author.}\\[.2cm]
  Zhejiang University, China
}

\begin{document}

\maketitle

\begin{abstract}
Long‐horizon video--audio reasoning and fine‐grained pixel understanding impose conflicting requirements on omnimodal models: dense temporal coverage demands many low-resolution frames, whereas precise grounding calls for high-resolution inputs.  We tackle this trade-off with a \emph{two-system architecture}: a \textbf{Global Reasoning System} selects informative keyframes and rewrites the task at low spatial cost, while a \textbf{Detail Understanding System} performs pixel-level grounding on the selected high-resolution snippets. 
 Because ``optimal'' keyframe selection and reformulation are ambiguous and hard to supervise, we formulate them as a reinforcement-learning (RL) problem and present \textbf{Omni-R1}—an end-to-end RL framework built on Group Relative Policy Optimization.  
 Omni-R1 trains the Global Reasoning System through hierarchical rewards obtained via online collaboration with the Detail Understanding System, requiring only one epoch of RL on small task splits.  
 Experiments on two challenging benchmarks, Referring Audio-Visual Segmentation (RefAVS) and Reasoning Video Object Segmentation (REVOS), show that Omni-R1 not only surpasses strong supervised baselines but also outperforms specialized state-of-the-art models, while substantially improving out-of-domain generalization and mitigating multimodal hallucination.  
 
 Our results demonstrate the first successful application of RL to large-scale omnimodal reasoning and highlight a scalable path toward universally foundation models. Our code is released at:
\url{https://github.com/aim-uofa/Omni-R1}.
\end{abstract}

\begin{figure}[!h]

    \centering
    \includegraphics[width=1.0\textwidth]{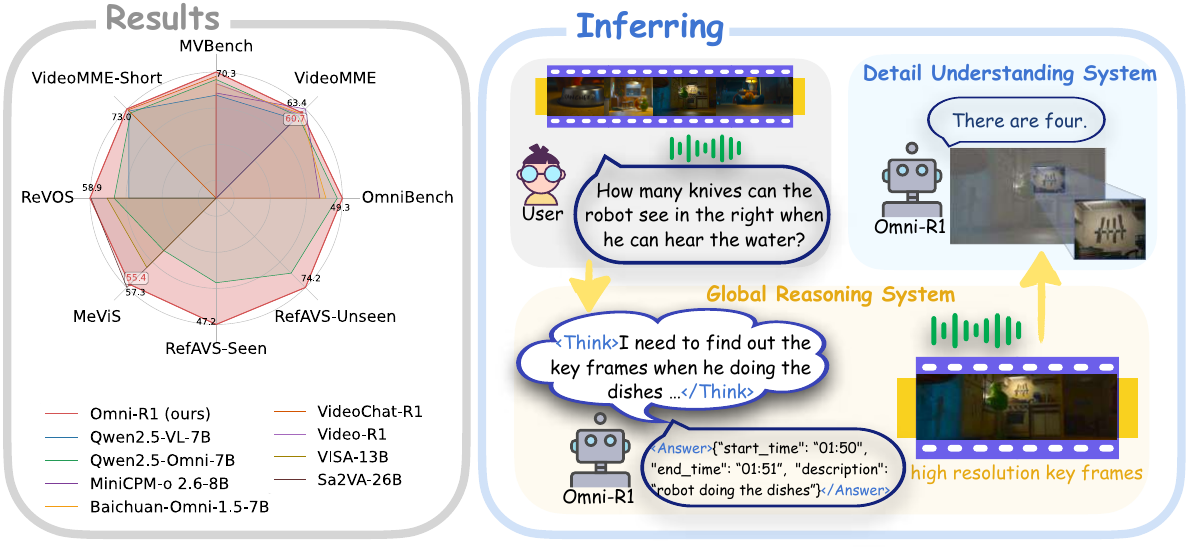}
    \includegraphics[width=1.0\textwidth]{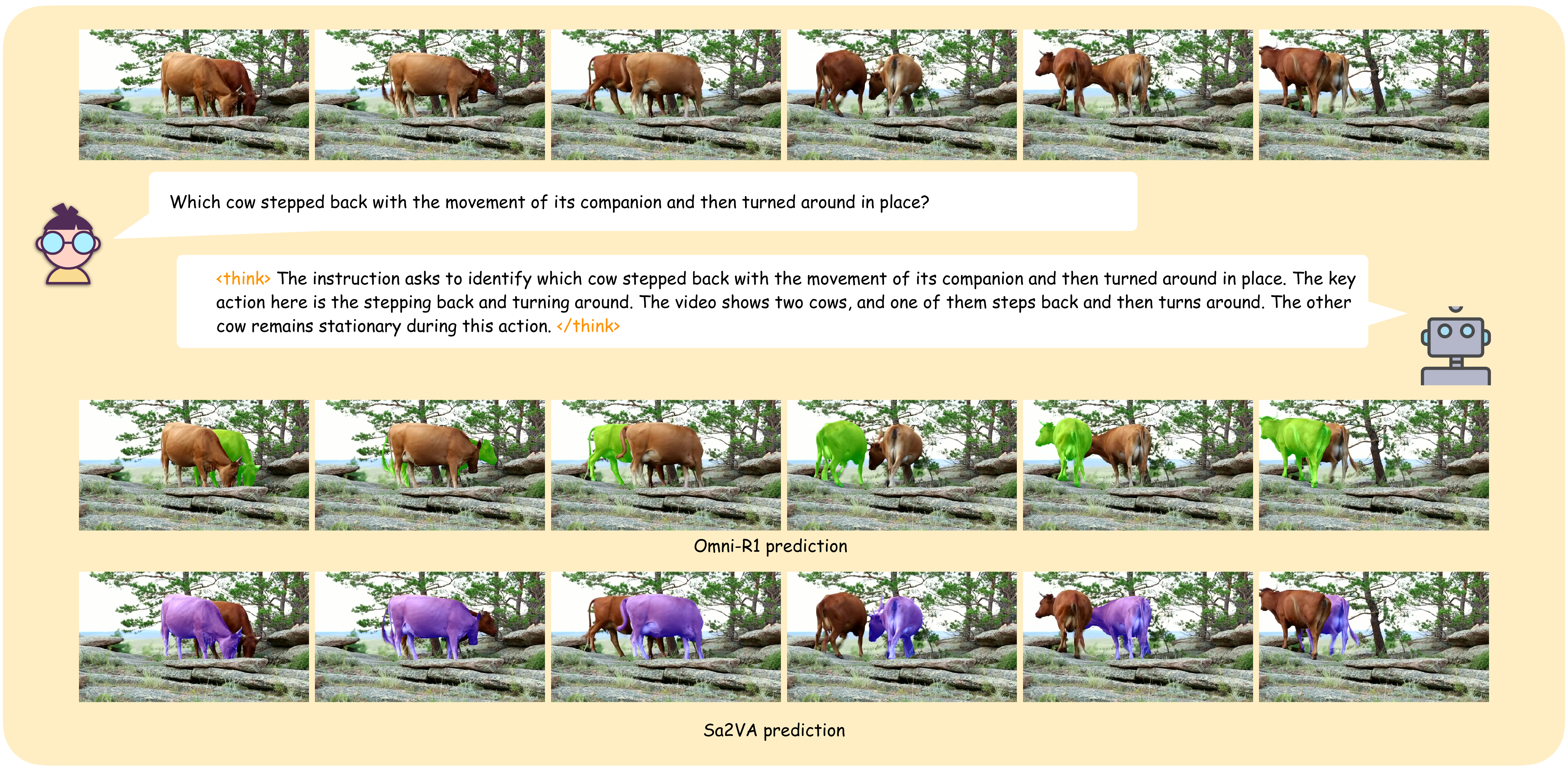}

    \caption{Overview of the proposed \textbf{Omni-R1} system for collaborative video understanding. Left: Performance comparison across multiple benchmarks shows Omni-R1 significantly outperforms existing omni-modal and video-reinforced MLLMs on both segmentation-centric and reasoning-centric tasks. Top-right: Omni-R1 employs a two-stage collaborative framework, integrating a detail understanding system (for precise visual QA) and a global reasoning system (for temporal grounding and high-resolution key frame identification). Bottom: A qualitative example highlights Omni-R1’s precise spatial-temporal segmentation and reasoning in identifying object-centric actions, outperforming prior expert models (e.g., Sa2VA) in complex scenarios.}
    \label{fig:main_fig}
\end{figure}

\section{Introduction}

Enabling models to simultaneously perceive, understand, and reason over omnimodal inputs—such as text, video, and audio—in complex real-world scenarios remains a longstanding goal in artificial intelligence  \cite{li2025baichuan, yao2024minicpm, hori2017attention,zellers2022merlot}.
Recent advances in omnimodal pretraining and instruction fine-tuning have led to the emergence of omnimodal models, bringing us closer to this objective  \cite{hurst2024gpt,team2024gemini,xu2025qwen25omni}.
Despite recent progress, current omnimodal models exhibit notable limitations in two key areas: (1) long-horizon reasoning over complex temporal sequences in video and audio, and (2) fine-grained spatial understanding at the pixel level.
A fundamental challenge underlying these two problems is the inherent trade-off between temporal coverage and spatial resolution  .
Long-horizon reasoning \cite{zou2024seconds,shi2025mavors} over video and audio typically requires high frame rates to capture global temporal context, which significantly increases memory and computational overhead—often forcing models to operate on low-resolution frames.
Conversely, fine-grained pixel understanding  \cite{lai2024lisa,bai2025qwen2} demands high-resolution inputs to preserve visual details, which in turn limits the number of frames that can be processed.
This trade-off creates a tension between global context modeling and local detail preservation, making it difficult for existing models to excel at both simultaneously.

A natural way to address this trade-off is to decompose the problem into two stages. 
Accordingly, we frame our solution as a two-system architecture:
\begin{itemize}
    \item 

System 1 (\textbf{Global Reasoning System}) performs coarse-grained, global reasoning over long video sequences at low spatial resolution—acting as a fast, context-aware selector that identifies critical temporal segments.
\item 
System 2 (\textbf{Detail Understanding System}), in contrast, conducts detailed, high-resolution analysis over a small number of keyframes, focusing on precise grounding and fine-grained understanding.

\end{itemize}

To illustrate how these systems interact, 
consider a task where the goal is to segment the last person to disappear (\textit{or} make a sound) in a scene.
\textbf{System1} first processes the full video (\textit{with} audio) sequence to determine, through low-resolution multimodal abstraction, which person is the last to leave visually or to emit sound. 
It then selects a few key segments where this individual appears or speaks.
Since \textbf{System 2} operates only on short segments with high-resolution input and lacks access to long-range temporal or auditory context, 
\textbf{System 1} needs to reformulate the original reference task—initially requiring long-horizon multimodal reasoning—into a simpler, localized problem.  
This reformulated task focuses on attributes, identity cues, and object permanence within the selected key segments, making it solvable using only fine-grained visual information.
\textbf{System 2} then takes these key segments and performs fine-grained visual grounding directly on the high-resolution input, bypassing the need for global reasoning.
This two-system design enables scalable and efficient multimodal reasoning by eliminating the need to process entire videos at high resolution, and effectively addresses the dual challenge of long-horizon reasoning and fine-grained visual understanding.

It is worth noting that current multimodal models already perform well as \textbf{Detail Understanding Systems} in tasks such as visual grounding \cite{bai2025qwen2,peng2023kosmos}, OCR \cite{yao2024minicpm,zhang2023llavar,liu2024textmonkey} and fine-grained image understanding \cite{rasheed2024glamm,xiao2024florence} on high-resolution inputs.  
Given this progress, the bottleneck in our two-system framework lies primarily in the capabilities of \textbf{System 1}.
In this work, we therefore focus on improving \textbf{Global Reasoning System}, particularly its ability to select informative keyframes and reformulate the task.
However, defining what constitutes an “optimal” keyframe selection or task reformulation is inherently ambiguous and task-dependent, making it impractical to rely on manually curated SFT data.
To address this, we propose \textbf{Omni-R1}, an end-to-end reinforcement learning framework tailored for omnimodal reasoning.  
Built upon the Group Relative Policy Optimization (GRPO) \cite{guo2025deepseekr1,shao2024deepseekmath} algorithm, our method simulates online collaboration between \textbf{System 1} and \textbf{System 2}, applying policy gradient updates guided by a hierarchical reward framework to progressively train System 1 to select keyframes and reformulate tasks in long-horizon, omnimodal settings.

From a reinforcement learning (RL) perspective, although it has proven effective in enhancing reasoning within large language models \cite{guo2025deepseekr1,shao2024deepseekmath,ouyang2022RLHF}, RL remains underexplored in omnimodal settings.
One major challenge lies in the lack of effective multimodal reasoning data \cite{wang2024omnibind}, along with uncertainty about whether language-based RL techniques can generalize across modalities.
While Omni-R1 bridges this gap by reformulating long-horizon multimodal understanding as a collaborative process between two systems.
In our design, the Global Reasoning System functions as an RL agent that selects keyframes and reformulates tasks for the Detail Understanding System to complete.
Such an approach provides a scalable path toward improving temporal reasoning and summarization in omnimodal models, while also opening new opportunities for applying RL beyond purely linguistic tasks.

To validate the effectiveness of Omni-R1, we benchmark it on two especially demanding tasks, namely Referring Audio-Visual Segmentation (RefAVS~\cite{wang2024ref}) and Reasoning Video Object Segmentation (REVOS~\cite{yan2024visa}),
both of which require temporal reasoning over 
video(audio) streams and fine-grained pixel understanding. 
Training Omni-R1 for just one epoch on the small datasets already lifts performance well beyond our baseline model and even surpasses the strongest, highly specialized state-of-the-art models on each benchmark.
Even more striking, reinforcement learning improves out-of-domain generalization, whereas conventional supervised fine-tuning often weakens it. Omni-R1 achieves higher scores in both pure video-understanding and omnimodal understanding settings, outperforming recent RL methods tailored specifically to video-reasoning tasks.
Finally, we conduct a comprehensive suite of diagnostic studies—including ablations over key architectural and training choices and an analysis of RL’s impact on multimodal hallucination—which together highlight the versatility and reliability of our approach. We hope 
that 
Omni-R1 offers a %
new 
direction for applying reinforcement learning to future all-modality foundation models.

Our primary contributions are summarized as follows:
\begin{itemize}
\itemsep 0cm
    \item   
          We present a scalable \textbf{Global Reasoning%
          }, and 
          {\bf Detail Understanding} two-system architecture
           that separates long-horizon video–audio reasoning from fine-grained pixel-level grounding, effectively resolving the temporal–spatial trade-off that constrains existing omnimodal models.
    \item  We introduce an end-to-end reinforcement-learning framework \textbf{Omni-R1}, built on Group Relative Policy Optimization that trains \textbf{System 1}—via hierarchical rewards and simulated collaboration with \textbf{System 2}—to select keyframes and reformulate tasks in long-horizon omnimodal settings.
    \item 
          With one epoch RL training, \textbf{Omni-R1} surpasses strong supervised baselines and specialized SOTA
          methods on RefAVS and REVOS, while markedly improving out-of-domain generalization including video understanding and omnimodal understanding.
\end{itemize}

\section{Related Work}
\label{sec:related_work}

\subsection{Omni-modal Large Models}
The advent of Large Language Models (LLMs) has revolutionized artificial intelligence, showcasing unprecedented capabilities in understanding, generating, and reasoning with textual data~\cite{achiam2023gpt, bai2023qwen, liu2024deepseekv3, touvron2023llama}. Building upon this foundation, Multimodal Large Language Models (MLLMs) have emerged, integrating multiple data modalities—such as vision, language, and audio—to achieve a more holistic understanding of complex tasks~\cite{bai2025qwen2, chen2024expanding, liu2023visual, liu2023llava, lu2024deepseekvl}. 

To differentiate from vision-language models (VLMs), multimodal large language models (MLLMs) incorporating the audio modality, such as Qwen2.5-Omni \cite{xu2025qwen2}, are termed \textit{omni-modal models}, abbreviated as \textit{omni}. MiniCPM-o 2.6~\cite{yao2024minicpm} extends its vision-language foundation~\cite{yao2024minicpm} with audio processing capabilities, allowing it to operate across more modalities. Baichuan-Omni-1.5~\cite{li2025baichuan}, trained and inferred in a fully end-to-end manner, surpasses GPT-4o-mini on the full-modality leaderboard OmniBench~\cite{li2024omnibench}. The recent development of omni-modal models further extends this integration, encompassing visual, linguistic, and auditory modalities to approach a comprehensive multimodal understanding~\cite{xu2025qwen25omni}.

\subsection{MLLM with RL}
Despite the remarkable progress enabled by supervised learning and instruction tuning, key challenges persist in aligning MLLMs with human preferences, mitigating harmful outputs, and enhancing their performance on complex reasoning tasks. Reinforcement Learning (RL), particularly Reinforcement Learning from Human Feedback (RLHF)~\cite{ouyang2022RLHF}, has proven effective in addressing these issues within unimodal LLMs, contributing to the success of models like ChatGPT~\cite{achiam2023gpt}.
A notable advancement in this domain is the introduction of DeepSeek-R1~\cite{guo2025deepseekr1}, which employs Group Relative Policy Optimization (GRPO) to enhance reasoning capabilities. GRPO innovatively replaces traditional critic models with a group-based reward normalization approach, reducing computational costs while maintaining performance~\cite{shao2024deepseekmath}. This technique has demonstrated that pure RL can effectively develop strong reasoning abilities without reliance on supervised data.

While reinforcement learning techniques have been widely explored in LLMs, their application to MLLMs is still at an early stage. Most recent efforts~\cite{liu2025visual,shen2025vlm,feng2025videor1,li2025videochat} have primarily focused on vision and language modalities, with little attention paid to more comprehensive multimodal integration. Notably, a concurrent work, R1-Omni~\cite{zhao2025r1omniexplainableomnimultimodalemotion}, is the first to include audio in addition to vision and language; however, its focus is limited to a single motion recognition task.

In contrast, our work targets more general long-horizon understanding tasks and conducts a more comprehensive and systematic investigation of omni-modal reinforcement learning. Building on recent advances, we propose \textbf{Omni-R1}, \textit{an omni-modal framework that unifies vision, language, and audio processing under an end-to-end RL optimization pipeline}. Our two-system design, which separates temporal reasoning from spatial perception, enables enhanced long-horizon understanding and fine-grained attention, allowing Omni-R1 to better address complex multimodal tasks requiring both structured perception and dynamic decision-making.

\section{Omni-R1}
\label{sec:omni-r1}

\subsection{Task and System Formulation}

We consider a long-horizon multimodal understanding task, where the model receives a video sequence $V = \{v_1, v_2, \dots, v_T\}$ and a synchronized audio stream $A = \{a_1, a_2, \dots, a_T\}$, along with an instruction or query $q$.
The goal is to produce a task-specific output $y$ (e.g., a localized segment, a textual response, or a grounding prediction) that reflects both global temporal reasoning and fine-grained visual understanding.
To better facilitate global temporal reasoning, we transform the raw instruction $q$ into a high-level instruction $q_{\text{global}} = \mathcal{T}(q)$ via a template-based rewriting function $\mathcal{T}$.

\paragraph{Stage 1: Global Reasoning System.}
We reduce the spatial resolution of the video as commonly adopted to obtain a low-resolution stream $\tilde{V} = \{\tilde{v}_1, \dots, \tilde{v}_T\}$ suitable for efficient global processing.  

Given $(\tilde{V}, A, q_{\text{global}})$, System 1 produces a set of $K$ selected segments(frames) $\mathcal{S} = \{s_1, s_2, \dots, s_K\}$ and a corresponding set of local queries $\{ q^{(i)}_{\text{local}} \}_{i=1}^K$
, intended to simplify the reasoning objective for System 2:
\[
\mathcal{S}, \{ q^{(i)}_{\text{local}} \}_{i=1}^K = \pi^{\text{(S1)}}(\tilde{V}, A, q_{\text{global}})
\]

\begin{figure}[t]

    \centering
    \includegraphics[page=2, width=\textwidth]{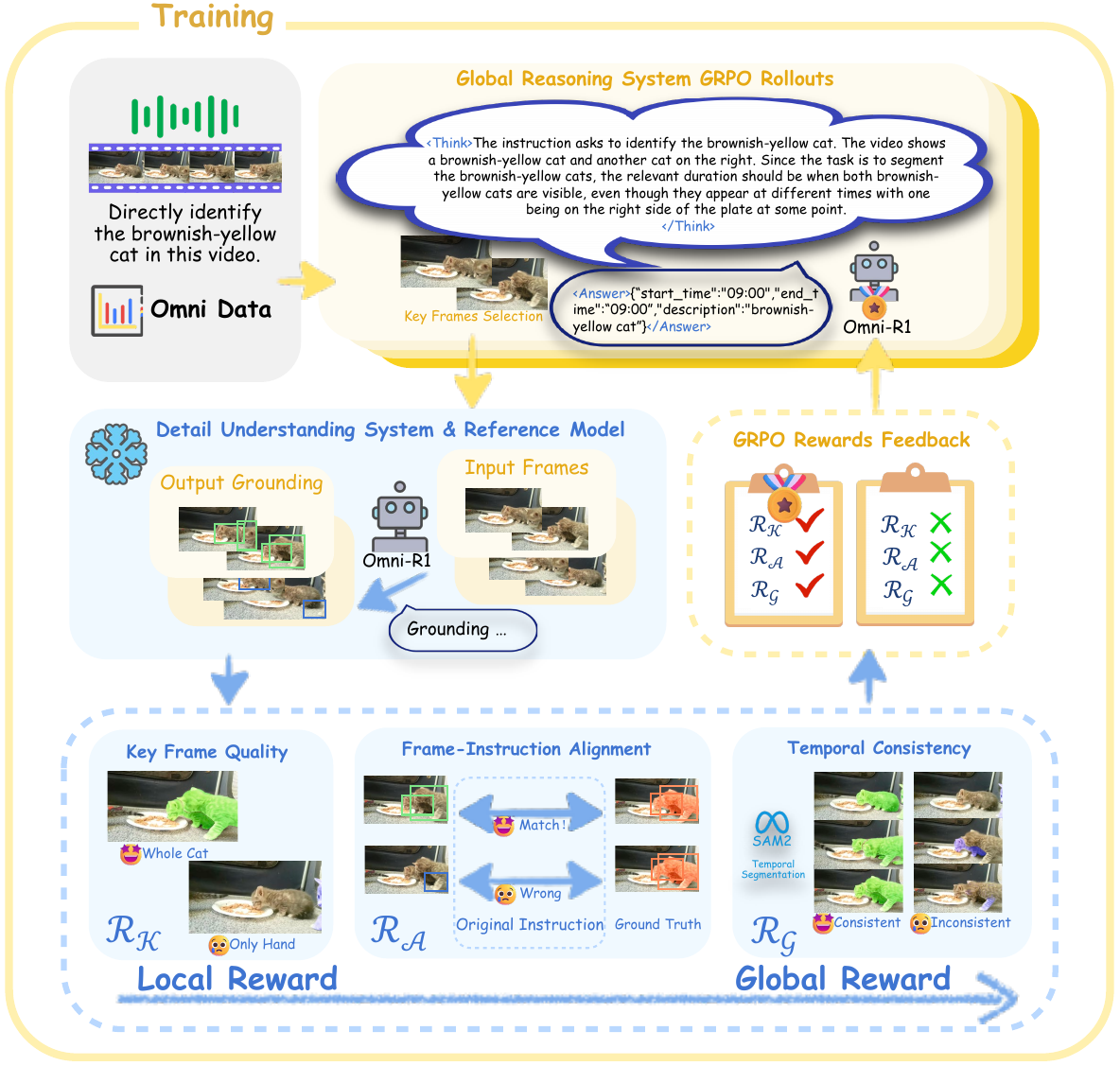}

    \caption{Exclusively trained as System 1 on video segmentation tasks in an End-to-End RL pipeline, Omni-R1 improved general understanding capabilities.}
    \label{fig:pipeline}
\end{figure}

\paragraph{Stage 2: Detail Understanding System.}
System 2 then receives the high-resolution frames $V_{\mathcal{S}} = \{v_{s_1}, \dots, v_{s_K}\}$ corresponding to the segments selected by System 1. 
Given $({V}_{\mathcal{S}}, \{ q^{(i)}_{\text{local}} \}_{i=1}^K)$, it performs fine-grained multimodal reasoning and produces the final output:
\[
y = \mathcal{\pi}^{\text{(S2)}}({V}_{\mathcal{S}}, \{ q^{(i)}_{\text{local}} \}_{i=1}^K)
\]

For tasks such as RefAVS and RVOS,  
one possible instantiation of System 2 is as a combination of a per-frame grounding model $\mathcal{F}_{\text{grounding}}$ and a frozen video segmentation model $\mathcal{F}_{\text{seg}}$ (e.g., SAM2 \cite{ravi2024sam}). Given the selected high-resolution frames $V_{\mathcal{S}} = \{v_{s_1}, \dots, v_{s_K}\}$ and the corresponding local instructions $\{q^{(i)}_{\text{local}}\}_{i=1}^K$, the grounding model is applied independently to each pair $(v_{s_i}, q^{(i)}_{\text{local}})$ to predict a set of bounding boxes $\mathcal{B}_{s_i} = \{ b^{(i)}_1, \dots, b^{(i)}_{N_i} \}$, where each $b^{(i)}_j \in \mathbb{R}^4$ denotes a box in $(x_1, y_1, x_2, y_2)$ format.
These predicted boxes are then passed to the segmentation model to produce pixel-level instance masks and propagate them temporally across the entire video: 
\[
\hat{\mathcal{M}} = \mathcal{F}_{\text{seg}}(V, V_{\mathcal{S}}, \{ \mathcal{B}_{s_i} \}_{i=1}^K)
\]
where the final output $\hat{\mathcal{M}} = \{\hat{m}_1, \dots, \hat{m}_T\}$ is a sequence of temporally aligned masks. The corresponding ground-truth mask sequence is denoted as $\mathcal{M}^* = \{m^*_1, \dots, m^*_T\}$, where each $m^*_t$ is the binary  segmentation mask for frame $v_t$.
For more details on the segmentation model $\mathcal{F}_{\text{seg}}$, please refer to Section~\ref{sec:appendix_a} in Appendix.

\subsection{End-to-End Reinforcement Learning via GRPO}
We now turn our focus to optimizing System 1 $\pi^{(\text{S1})}$, with the goal of improving both the selection of key segments $\mathcal{S}$ and the formulation of task-specific local instructions $\{q^{(i)}_{\text{local}}\}$, in order to better support System 2 in performing fine-grained understanding. 
However, $\mathcal{S}$, $\{q^{(i)}_{\text{local}}\}$, and the System 2 (i.e., $\pi^{(\text{S2})}$) are strongly coupled, making it difficult to directly define what constitutes an optimal pair $(\mathcal{S}, \{q^{(i)}_{\text{local}}\})$ for downstream performance. 
As a result, constructing high-quality supervised fine-tuning (SFT) data for $\pi^{(\text{S1})}$ is infeasible.

Instead, we propose to optimize $\pi^{(\text{S1})}$ via reinforcement learning by designing a reward function $R(\mathcal{S}, \{q^{(i)}_{\text{local}}\}, \pi^{(\text{S2})})$ that evaluates the utility of System 1’s outputs in enabling System 2 to succeed. Under this framework, $\pi^{(\text{S1})}$ is trained to explore and generate candidate outputs, and receives feedback from the environment through this reward.
Specifically, we adopt a GRPO-based policy optimization scheme. At each iteration, we sample $N$ responses from the current policy $\pi^{(\text{S1})}$ and compute the corresponding rewards $r_n$ using the reward function $R(\cdot)$. We then normalize the rewards to estimate the advantage of each sample:
\begin{equation}
A_n = \frac{r_n - \text{mean}(\{r_1, \ldots, r_N\})}{\text{std}(\{r_1, \ldots, r_N\})}
\label{eq:advantage_n_modified}
\end{equation}
Based on the computed advantages $\{A_n\}$, we perform PPO-style policy gradient updates to improve $\pi^{(\text{S1})}$.

\subsection{Hierarchical Reward Design for System 1}

Designing an effective reward function that accurately reflects the quality of the action pair $(\mathcal{S}, \{q^{(i)}_{\text{local}}\})$ and provides a meaningful training signal for System 1 ($\pi^{(\text{S1})}$) is critical to the success of our framework. 
In this section, we describe our hierarchical reward formulation tailored for the Referring Video Object Segmentation (RVOS) task, which aims to guide System 1 to progressively learn to select informative keyframes and generate useful local instructions.

Due to the strong coupling among $\mathcal{S}$, $\{q^{(i)}_{\text{local}}\}$, and System 2 ($\pi^{(\text{S2})}$), relying solely on the final task objective (e.g., segmentation mIoU) as the reward leads to unstable and inefficient training. This is because such reward signals are sparse, non-decomposable, and difficult to attribute back to specific decisions made by $\pi^{(\text{S1})}$.
To address this, we propose a set of hierarchical reward functions, organized from weakly coupled to strongly coupled, and from local to global. These rewards are designed to incrementally shape the learning of System 1, starting from simpler supervision signals and gradually incorporating a more task-specific structure. We define three types of rewards:

\textbf{Key Frame Quality Reward} ($R_{\mathcal{K}}$): This reward evaluates the quality of the selected keyframes $\mathcal{S}$, independently of the instructions or the performance of subsequent segmentation.

It provides early learning signals to encourage the selection of visually salient or semantically diverse frames.

We define the Key Frame Quality Reward as a weighted combination of three factors:

\[
R_{\mathcal{K}} = \lambda_1 R_{\text{diversity}} + \lambda_2 R_{\text{num}} + \lambda_3 R_{\text{saliency}}
\]

The first term, \textit{Temporal Diversity Reward} $R_{\text{diversity}}$, encourages selected frames to spread over the video timeline, rather than being clustered within a short segment. 
This promotes broader temporal coverage and helps the model focus on long-range dynamics.

The second term, \textit{Frame Count Regularization} $R_{\text{num}}$, regularizes the number of selected frames $K$ to stay near a predefined target $K_0$, penalizes selections that include either too few or too many frames.

The third term, \textit{Object-Centric Saliency Reward} $R_{\text{saliency}}$, rewards keyframes that contain a large visible portion of the target object. 
This is based on the hypothesis that selecting such frames provides stronger visual anchors, which can facilitate more accurate and stable object tracking and segmentation throughout the video.
It is calculated as the normalized average GT mask area:
\[
R_{\text{saliency}} = \frac{1}{K} \sum_{i=1}^{K} \frac{\text{area}(m^*_{s_i})}{\max_{t} \text{area}(m^*_t)}
\]

Together, these components guide System 1 to select keyframes that are temporally diverse, reasonably sparse, and visually informative. Formal definitions of the reward are provided in Appendix Section~\ref{sec:appendix_a}.

\textbf{Frame-Instruction Alignment Reward} ($R_{\mathcal{A}}$) measures how well each local instruction $q^{(i)}_{\text{local}}$ aligns with its corresponding keyframe $v_{s_i}$. This reward evaluates whether the instruction provides sufficient grounding cues to locate the correct object in the frame. As it operates independently per frame-instruction pair, it does not depend on the segmentation model $\mathcal{F}_{\text{seg}}$, and thus ignores temporal consistency.
Concretely, given a frame $v_{s_i}$ and its corresponding instruction $q^{(i)}_{\text{local}}$, we apply the grounding model $\mathcal{F}_{\text{grounding}}$ to predict a set of bounding boxes $\mathcal{B}_{s_i} = \{ b^{(i)}_1, \dots, b^{(i)}_{N_i} \}$. 
We compare these predictions against the ground-truth target boxes $\mathcal{B}^*_{s_i}$ defined for that frame. 
Since a single instruction may refer to multiple target objects, both $\mathcal{B}_{s_i}$ and $\mathcal{B}^*_{s_i}$ can contain multiple instances.The reward is computed as the negative Hungarian matching loss commonly used in object detection \cite{carion2020end}:

\begin{equation}
R_{\mathcal{A}} = \frac{1}{K} \sum_{i=1}^{K} \left( 1 - \mathcal{L}_{\text{Hungarian}}(\mathcal{B}_{s_i}, \mathcal{B}^*_{s_i}) \right)
\end{equation}

This loss is minimized when the predicted boxes exactly match the ground-truth targets.

\textbf{Global Temporal Consistency Reward} ($R_{\mathcal{G}}$) is the most strongly coupled and task-specific reward in our framework, directly reflecting the final objective of long-term video object segmentation. Unlike previous rewards, which evaluate the selected keyframes or instructions in isolation, $R_{\mathcal{G}}$ jointly considers how the selected keyframes $\mathcal{S}$ and local instructions $\{q^{(i)}_{\text{local}}\}$ influence the performance of System 2 throughout the video.
This reward is designed to capture both the spatial accuracy and the temporal consistency of the predicted instance masks. 
In particular, it encourages System 1 to select frames that are critical for robust tracking—such as those appearing after significant object deformations, occlusions, or disappearances—so that the segmentation model (e.g., SAM2) can re-anchor to the target effectively.
Formally, given a candidate keyframe set $\mathcal{S}$ and corresponding instructions $\{q^{(i)}_{\text{local}}\}$, we feed them into System 2 ($\pi^{(\text{S2})}$) to obtain a full sequence of predicted masks $\hat{\mathcal{M}} = \{\hat{m}_1, \dots, \hat{m}_T\}$. 

The reward is computed as the average frame-wise Intersection over Union (IoU) with the ground-truth masks $\mathcal{M}^* = \{m^*_1, \dots, m^*_T\}$:

\begin{equation}
R_{\mathcal{G}} = \frac{1}{T} \sum_{t=1}^{T} \text{IoU}(\hat{m}_t, m^*_t)
\end{equation}

Finally, we combine the above three components to form the overall reward used for training System 1. The total reward is a weighted sum of the three terms:

\begin{equation}
R = \alpha_{\mathcal{K}} R_{\mathcal{K}} + \alpha_{\mathcal{A}} R_{\mathcal{A}} + \alpha_{\mathcal{G}} R_{\mathcal{G}}
\end{equation}

where $\alpha_{\mathcal{K,A,R}}$ are the weighting coefficients that control the importance of each reward component.

\section{Experiments}
\label{sec:experiments}
\subsection{Experiments Setting}
\label{sec:exp_setting}

\paragraph{System 1 and System 2}

We adopt \textbf{Qwen2.5-Omni-7B}~\cite{xu2025qwen25omni} as our base model, which serves as \textbf{System 1} responsible for high-level reasoning. To construct a lightweight and stable \textbf{System 2} during training, we use a frozen copy of the same pretrained Qwen2.5-Omni model, which also functions as a reference policy model for guiding optimization. For evaluation, unless otherwise stated, \textbf{Omni-R1} is serving as both System 1 and System 2 for resource efficiency. However, due to the modular design and decoupled functionality of the two systems, System 2 can be flexibly replaced with a stronger perception module in a \textbf{zero-shot} manner.

\paragraph{Training Paradigm}

We train System 1 on 1,600 samples randomly selected from the RefAVS~\cite{wang2024ref} dataset and 2,600 videos from the ReVOS~\cite{yan2024visa} and MeViS~\cite{MeViS} datasets for 1 epoch. To further enhance the model’s fine-grained understanding capabilities as system 2, we additionally train the model on 2,000 images from refCOCOg~\cite{mao2016generation} for one epoch in the style of SegZero~\cite{liu2025segzero}. Unless otherwise specified, all experiments are conducted using a policy KL divergence hyperparameter of $\beta = 0.04$, a group size of 8, and an initial learning rate of $1 \times 10^{-6}$ under the AdamW optimizer with a weight decay of 0.01. We adopt \texttt{sam2-hiera-large} as our SAM2~\cite{ravi2024sam} version throughout the experiments.

\newcommand{\nj}{$\mathcal{J}$}
\newcommand{\nf}{$\mathcal{F}$}
\newcommand{\njf}{$\mathcal{J}\&\mathcal{F}$}

\subsection{Referring Video Segmentation}

\paragraph{Referring Audio-Visual Segmentation} Ref-AVS~\cite{wang2024ref} is specifically designed for audio-visual segmentation tasks, offering a diverse and well-annotated collection of samples that require integrated reasoning across both modalities. The dataset comprises 2,908 audio-equipped video clips in the training set, covering 5,366 annotated objects across 39 semantic categories.

\begin{table}[t]
\centering
\caption{Performance comparison across models grouped by Seen and Unseen sets in Ref-AVSBench~\cite{wang2024ref}. Some metrics curated from ~\cite{wang2024ref}. \njf\ represents the average of (\nj) score and (\nf) score. $\dag$ indicates the results are tested on the masks predicted by SAM2 according to model's grounding output.}
\label{tab:grouped-performance}
\begin{tabular}{l *{6}{S[table-format=+2.1]}}
\toprule
\multirow{2}{*}{\textbf{Model}} 
& \multicolumn{3}{c}{\textbf{Seen}}
& \multicolumn{3}{c}{\textbf{Unseen}} \\
& \njf & \nj & \nf & \njf & \nj & \nf \\
\midrule
AVSBench~\cite{zhouavsbench} + \texttt{text} 
& \num{37.2} & \num{23.2} & \num{51.1} & \num{43.5} & \num{32.4} & \num{54.7} \\
AVSegFormer~\cite{gao2024avsegformer} + \texttt{text} 
& \num{40.2} & \num{33.5} & \num{47.0} & \num{43.1} & \num{36.1} & \num{50.1} \\
GAVS~\cite{wang2024prompting} + \texttt{text} 
& \num{39.4} & \num{28.9} & \num{49.8} & \num{39.8} & \num{29.8} & \num{49.7} \\
ReferFormer~\cite{wu2022language} + \texttt{audio} 
& \num{40.7} & \num{31.3} & \num{50.1} & \num{39.6} & \num{30.4} & \num{48.8} \\
R2VOS~\cite{li2023robust} + \texttt{audio} 
& \num{33.0} & \num{25.0} & \num{41.0} & \num{38.9} & \num{27.9} & \num{49.8} \\
EEMC~\cite{wang2024ref} 
& \num{42.8} & \num{34.2} & \num{51.3} & \num{57.2} & \num{49.5} & \num{64.8} \\
$\text{Qwen2.5-Omni-7B}^{\dag}$ 
& \num{31.6} & \num{27.7} & \num{35.5} & \num{62.3} & \num{59.0} & \num{65.7} \\
\midrule
$\text{Qwen2.5-Omni-7B}^{\dag}$(SFT) 
& \num{39.1} & \num{35.4} & \num{42.8} & \num{66.2} & \num{63.1} & \num{69.3} \\
$\text{Omni-R1-7B}^{\dag}$ 
& \bfseries\num{47.2} & \bfseries \num{43.0} & \bfseries \num{51.4} 
& \bfseries\num{74.2} & \bfseries \num{71.3} & \bfseries \num{77.0} \\
$\Delta$
& \bfseries \num{+16.4} & \bfseries \num{+15.3} & \bfseries \num{+9.4}
& \bfseries \num{+8.0} & \bfseries \num{+8.2} & \bfseries \num{+7.7} \\
\bottomrule
\end{tabular}
\end{table}

We evaluated the performance of our collaborative system on Ref-AVSBench~\cite{wang2024ref} with other Referring AVS methods. Omni-R1 outperforms previous SOTA EMMC~\cite{wang2024ref} by \textbf{+4.6\%} on \njf\ in seen set and \textbf{+17.0\%} on unseen set.

\paragraph{Reasoning Video Object Segmentation}
ReVOS~\cite{yan2024visa} is a VOS dataset that emphasizes reasoning about temporal behaviors through implicit object descriptions, comprising 35,074 pairs of instruction-mask sequences derived from 1,042 diverse videos. In contrast to traditional referring video segmentation datasets, ReVOS includes text instructions that necessitate a sophisticated understanding of both video content and general world knowledge.
% whereas MeViS centers on dynamic visual segmentation with a strong emphasis on comprehensive video content understanding. 

For our evaluation, we exclusively employed Sa2VA as \textbf{System 2} to investigate the full reasoning capabilities of Omni-R1 as \textbf{System 1}.

\begin{table}[bt]
\centering
\caption{Reasoning Video Object Segmentation performance comparison across different methods, the metric is \njf\ score(\%). $\ddag$ means the results are evaluated where Omni-R1-7B serves as System \num{1} and Sa2VA as System \num{2}(\num{1}B and \num{4}B).}
\label{tab:vos-comparison}
\begin{tabular}{l *{6}{S}}
\toprule
\multirow{2}{*}{\textbf{Model}} & \multicolumn{5}{c}{\textbf{ReVOS}} \\ %& \textbf{MeViS} \\

& \textit{Referring} & \textit{Reasoning} & \textit{Single} & \textit{Multi} & \textbf{\textit{Overall}} \\ %& \textbf{\textit{Average}} \\
\midrule
LISA-13B~\cite{lai2024lisa} & - & - & - & - & \num{41.6} \\ %& - \\
TrackGPT-13B~\cite{stroh2024trackgpt}  & - & - & - & - & \num{45.0} \\ %& - \\
VISA-13B~\cite{yan2024visa}  & - & - & - & - & \num{50.9} \\ %& \num{44.5} \\
Sa2VA-8B~\cite{yuan2025sa2va} & - & - & - & - & \num{57.6} \\ %& \num{46.9} \\
Sa2VA-26B~\cite{yuan2025sa2va} & - & - & - & - & \num{58.4} \\ %& \num{46.2} \\
\midrule
$\text{Qwen2.5-Omni-7B}^{\dag}$ & \num{46.3} & \num{26.9} & \num{38.6} & \num{37.4} & \num{36.6} \\ %& \num{33.6} \\
$\text{Omni-R1-7B}^{\dag}$ & \num{53.2} & \num{41.9} & \num{48.3} & \num{46.5} & \num{47.6} \\ %& \num{34.9} \\
$\text{Omni-R1-8B}^{\ddag}$ & \num{61.6} & \num{50.7} & \num{56.6} & \num{47.3} & \num{56.2} \\ %&  \bfseries \num{55.4} \\
$\text{Omni-R1-11B}^{\ddag}$ & \bfseries \num{64.1} & \bfseries \num{53.7} & \bfseries \num{59.2} & \bfseries \num{51.0} & \bfseries \num{58.9} \\ %& \num{54.6} \\
\bottomrule
\end{tabular}
\end{table}

Our \textbf{System 1} exhibits strong performance on video object segmentation tasks under both basic and reasoning-intensive conditions. When deployed as both systems ($\dagger$), Omni-R1-7B significantly outperforms the base model on ReVOS, achieving a \textbf{+11.0\%} improvement over Qwen2.5-Omni-7B. This result underscores its enhanced temporal reasoning and fine-grained recaption capabilities.

Furthermore, the collaborative system ($\ddag$) Omni-R1-11B achieves a score of \textbf{58.9\%} on ReVOS, surpassing much larger segmentation-specialized models such as Sa2VA-26B~\cite{yuan2025sa2va}. Notably, it achieves the best performance across all categories, including the reasoning subset in ReVOS (53.7\%), underscoring the effectiveness of our disentangled system architecture and reinforcement learning-based training paradigm.

\subsection{General Omni-Modal Understanding}

In this section, we focus on the impressive progress of Omni-R1 on multi-modal tasks, in comparison to its base model Qwen2.5-Omni-7B and other leading multi-modal models.

\setlength{\tabcolsep}{10pt} %
\begin{table}[b]
\centering
\caption{Performance comparison across models on general understanding QA benchmarks Omnibench, VideoMME, and MVBench.}
\label{tab:general-comparison}
\begin{tabular}{lcccccc}
\toprule
\multirow{2}{*}{\textbf{Method}} & \textbf{Omnibench} & \multicolumn{2}{c}{\textbf{VideoMME}} & \textbf{MVBench} \\
 & \textbf{Avg} & \textbf{General} & \textbf{Short} & \textbf{General} \\
\midrule
\multicolumn{5}{c}{\small{\textit{Vision-Language Models}}} \\
\midrule
Qwen2.5-VL-7B(CoT) & - & 56.1 & 71.3 & 57.4 \\
LLaVA-OneVision-7B~\cite{li2024llavaov} & - & 58.2 & - & 56.7 \\
Kangeroo-8B~\cite{kangaroogroup} & - & 56.0 & - & 61.1 \\
VideoChat-R1~\cite{li2025videochat} & - & - & 72.2 & 67.9 \\
Video-R1~\cite{feng2025videor1} & - & 59.3 & - & 63.9 \\
Sa2VA-26B~\cite{yuan2025sa2va} & - & 52.6 & - & - \\
\midrule
\multicolumn{5}{c}{\small{\textit{Omni-Modal Language Models}}} \\
\midrule
VITA-1.5-7B~\cite{fu2025vita} & 33.4 & 57.3 & - & 55.5 \\
MiniCPM-o 2.6-7B~\cite{yao2024minicpm} & 40.5 & 63.4 & - & 58.6 \\
Baichuan-Omni-1.5-7B~\cite{li2025baichuan} & 42.9 & 60.1 & - & 63.7 \\
Qwen2.5-Omni-7B & 47.3 & 58.3 & 69.8 & 66.1 \\
Omni-R1-7B & \bfseries 49.3  & \bfseries 60.7  & \bfseries 73.0  & \bfseries 70.3 \\
$\Delta$ & \bfseries \num{+2.0} & \bfseries \num{+2.4} & \bfseries \num{+3.2} & \bfseries \num{+4.2} \\
\bottomrule
\end{tabular}
\end{table}

Omni-R1 shows stable improvements over its base model Qwen2.5-Omni. Omni-R1 achieves an average improvement of \textbf{+2.0\%}, \textbf{+2.7\%} and \textbf{+3.7\%} over baseline on OmniBench~\cite{li2024omnibench}, VideoMME~\cite{fu2024videomme} and MVBench~\cite{li2024mvbench} respectively, surpassing all other open-source omni-models. 
Specifically, in video understanding tasks, Omni-R1 gains more progress on short videos (less than 2 min) than long videos. This could be attributed to our VOS training videos, where almost all videos are less than 2 minutes, with MeViS even being less than 30 seconds.

\paragraph{System 1's Strength in General Understanding Tasks}
Omni-R1 demonstrates significant improvements, achieving outstanding general performance on the omni-modal benchmark OmniBench, where it outperforms all other 7B models in the open-source space. With a score of \textbf{72.5} in the short subset of VideoMME, Omni-R1 surpasses VideoChat-R1~\cite{li2025videochat}, which was exclusively fine-tuned for Video QA tasks through RL. Additionally, Omni-R1 achieves the highest score on MVBench, outperforming all other omni-modal models by a large margin.

The strong performance of Omni-R1 across both in-domain and general tasks showcases the effectiveness of our reinforcement learning approach. Leveraging System 1's multi-modal reasoning, the model excels in task-specific scenarios and generalizes effectively to unseen tasks, demonstrating its robustness and adaptability in real-world environments.

\section{Conclusion}
\label{sec:conclusion}
We present Omni-R1, a novel reinforcement learning framework that addresses a key limitation in omnimodal models: the trade-off between long-horizon temporal reasoning and fine-grained spatial understanding. By decoupling these objectives into a two-system architecture comprising a Global Reasoning System and a Detail Understanding System Omni-R1 enables scalable and efficient processing of complex video–audio–text inputs.

Through task reformulation and keyframe selection trained via Group Relative Policy Optimization, our approach significantly improves performance on challenging benchmarks like RefAVS and ReVOS, while also enhancing out-of-domain generalization. Our diagnostic studies further confirm the robustness and versatility of the framework. We hope that  this work opens new avenues for integrating reinforcement learning into next-generation omnimodal foundation models.

\clearpage
\appendix
\section{Appendix Overview}

This appendix provides additional details on the experimental setup, model architecture along with training pipeline, and supplementary results that support the findings presented in the main paper.

\begin{itemize}
    \item \textbf{\nameref{sec:appendix_a}:} this section details additional aspects of our method: our two-system architecture (including user instructions for its MLLM components and prompts for the downstream SAM2 model), our reward design, and the differences between training and inference procedures.
    \item \textbf{\nameref{sec:appendix_b}:} this section provides ablation studies of System 1 on the reward component and dataset selection.
    \item \textbf{\nameref{sec:appendix_c}:} this section provides more visualization results, including examples in comparison with other methods and failure case analysis.
    \item \textbf{\nameref{sec:appendix_d}:} this section provides analysis on the hallucination issue and the influence of resolution on general video understanding tasks.
    \item \textbf{\nameref{sec:appendix_e}:} this section discusses the limitations of our method and potential future work.
\end{itemize}

\section{Implementation Details}
\label{sec:appendix_a}

\paragraph{User Instructions on Two Systems.}

To enable MLLMs to perform keyframe selection and referred object captioning, we designed the prompt as shown in the figure \ref{fig:sys1-prompt-vos}. We formulate keyframes as time duration segments and assign spatial description text to each duration. Additionally, we observed that during training, the model could be influenced by the timestamp patterns seen in the prompt examples. Therefore to increase the diversity of keyframe distributions during training and prevent the model from overfitting to specific timestamps, we incorporated randomized timestamps into the prompt, encouraging the model to focus on learning keyframe selection and caption rather than simply copying training timestamps. For AVS tasks, we designed a similar prompt (see Figure \ref{fig:sys1-prompt-avs}) to guide the model in analyzing the audio content and identifying the corresponding visual grounding description. The prompt emphasizes the need to avoid temporal expressions and instead focus on visual cues.

The intermediate results are then interpreted as frames and paired descriptions before being fed into System 2. The prompt for System 2 follows the official grounding prompt of Qwen2.5-VL, where the output is a list of bounding boxes and their corresponding labels in JSON format.

\paragraph{Prompt Design for SAM2 as Downstream Segmenter.}
Once the keyframe grounding results are obtained from System 2, we assign a unique identifier to each detection result using a tuple format: \texttt{(roll\_out\_idx, frame\_idx, pred\_obj\_idx, bbox)}. This ensures that any detection within a single GRPO group can be uniquely referenced.

Since all detections within a group share the same input context, we optimize inference efficiency by processing all detection results in a single forward pass. Specifically, we feed the entire video segment into SAM2, and for each detection tuple, we assign a unique object ID. These object IDs are used as conditioning inputs to SAM2 to obtain their respective segmentation masks.

Specifically, we maintain a mapping dictionary between detection tuples and assigned object IDs $\mathcal{P}: \texttt{tuple} \rightarrow \texttt{obj\_id}$, which enables us to reverse-map SAM2's outputs $\{\hat{\mathcal{M}}_{\texttt{obj\_id}}| \texttt{obj\_id} \in \mathcal{P}(\texttt{tuple})\}$ back to the original detection structure $\hat{\mathcal{M}}_{\texttt{tuple}}$. The segmentation results are then matched to the corresponding grounded predictions and used for reward evaluation.

\paragraph{Reward Design of $R_{\mathcal{K}}$}

To enable the model to learn from diverse keyframe selections, we design an evaluation reward function, $R_{\mathcal{K}}$, which assesses both the diversity and quality of the chosen keyframes. This function is formulated as:
$$R_{\mathcal{K}} = \lambda_1 R_{\text{diversity}} + \lambda_2 R_{\text{num}} + \lambda_3 R_{\text{saliency}}$$
where the last component, $R_{saliency}$, has been clarified in the main paper. Here, we detail the first two reward components.

Firstly, to discourage the model from selecting temporally adjacent keyframes, which can lead to redundant System 2 inference, we introduce a distribution reward function, $R_{diversity}$. This component evaluates the distributional diversity of keyframes by calculating the temporal intervals between them. Specifically, all keyframes are sorted in chronological order. We then compute the temporal interval $t_{i+1} - t_i$ between each pair of consecutive keyframes. The final $R_{diversity}$ value is subsequently determined based on the collection of all such inter-frame temporal intervals.

The diversity reward $\mathcal{R}_{\text{diversity}}(\mathcal{S})$ can be defined as the sum of an overlap punishment term and a distribution reward term:
\[
    \mathcal{R}_{\text{diversity}}(\mathcal{S}) = \textit{overlap\_punish} \cdot |\mathcal{I}| + \textit{dist\_reward} \cdot |\mathcal{D}|
\]
where:
\begin{itemize}
    \item $\mathcal{S}$ is the set of selected items. Let $\mathcal{S}_{\text{sorted}} = (s_1, s_2, \ldots, s_M)$ be the sequence of $M=|\mathcal{S}|$ items sorted according to the relevant criteria (e.g., timestamps).
    \item $\mathcal{I}$ is the set of indices $i$ for which an "overlap" condition is met between $s_i$ and $s_{i+1}$. Specifically, assuming $\mathrm{idx}(s_j)$ gives an identifier for item $s_j$:
          \[
              \mathcal{I} = \{i \in \{1, \ldots, M-1\} \mid \mathrm{idx}(s_i) = \mathrm{idx}(s_{i+1})\}
          \]
          $|\mathcal{I}|$ is the number of such identified overlaps (e.g., pairs of consecutive items with identical identifiers).
    \item $\mathcal{D}$ is the set of indices of items in $\mathcal{S}_{\text{sorted}}$ that are not considered the start of an overlap as defined by $\mathcal{I}$:
          \[
              \mathcal{D} = \{j \in \{1, \ldots, M\} \mid j \notin \mathcal{I}\}
          \]
          Therefore, $|\mathcal{D}| = M - |\mathcal{I}|$.
    \item $\textit{overlap\_punish}$ is the coefficient for the punishment. For this term to act as a punishment, $\textit{overlap\_punish}$ should typically be a negative value (e.g., $-0.2$), or if it's a positive value, it should be subtracted from the reward.
    \item $\textit{dist\_reward}$ is the coefficient for the reward given to items not initiating an overlap.
\end{itemize}

The formula $\mathcal{R}_{\text{diversity}}(\mathcal{S})$ can also be written as:
\[
    \mathcal{R}_{\text{diversity}}(\mathcal{S}) = (\textit{overlap\_punish} - \textit{dist\_reward}) \cdot |\mathcal{I}| + \textit{dist\_reward} \cdot M
\]

\paragraph{Reward Design of $R_{\mathcal{A}}$}

The specific formulation of $R_{\mathcal{A}}$ is as follows:
\[
    R_{\mathcal{A}} = \frac{1}{K} \sum_{i=1}^{K} \left( 1 - \mathcal{L}_{\text{Hungarian}}(\mathcal{B}_{s_i}, \mathcal{B}^*_{s_i}) \right)
\]

where $\mathcal{B}_{s_i}$ denotes the set of predicted bounding boxes at the $s_i$-th frame, and $\mathcal{B}^*_{s_i}$ represents the corresponding set of ground truth bounding boxes. The function $\mathcal{L}_{\text{Hungarian}}$ refers to the Hungarian matching loss~\cite{carion2020end,7738348}, and $K$ is the total number of selected keyframes.

The Hungarian matching loss $\mathcal{L}_{\text{Hungarian}}$ is computed based on the Intersection-over-Union (IoU) between predicted and ground truth bounding boxes. Specifically, a cost matrix $\mathcal{M}$ is first constructed using the IoU values between each pair of predicted and ground truth boxes. Then, the Hungarian algorithm is applied to the negative matrix $-\mathcal{M}$ to obtain the optimal one-to-one matching that minimizes the total negative cost, which corresponds to maximizing the overall IoU-based matching accuracy.

\paragraph{Reward Design of $R_{\mathcal{G}}$}
For $R_{\mathcal{G}}$, we adopt a simple aggregated IoU as the reward function.
Specifically, for each detected object, we accumulate the predicted segmentation masks across all objects to construct a per-frame mask set $\hat{\mathcal{M}}_{t}$. Then, we compute the Intersection-over-Union (IoU) between the predicted masks and the corresponding ground truth masks $\mathcal{M}^{*}_{t}$  on each frame. The final reward is obtained by averaging the IoU values across all frames.
\[
    R_{\mathcal{G}} = \frac{1}{T} \sum_{t=1}^{T} \text{IoU}(\hat{m}_t, m^*_t)
\]

\paragraph{Training and Inference Strategy}
For video clips, we first feed them into System 1 at a relatively low resolution of a per-frame pixel \texttt{128$\times$28$\times$28}, which allows us to process longer video segments during training and inference. Then System 2 predicts detection results at a higher resolution of \texttt{900$\times$28$\times$28}.
\textbf{For VOS tasks}, we adopt a random uniform sampling strategy during training, selecting between 8 and 24 frames per video to enhance temporal diversity and robustness. All SAM2-based segmentation and reward evaluations are then applied to these resampled clips at their original input resolution. \textbf{For RefAVS tasks}, we observed severe cross-modal hallucination issues during preliminary experiments, particularly when reasoning jointly over full-length audio and multi-frame video inputs. To mitigate this, we introduce a simplified variant, RefAID (Referring Audio-Image Detection), which reduces the AVS problem to object detection using only the first video frame and the corresponding full audio query. In this setting, no SAM2 segmentation is used; training is driven solely by detection-based rewards.

During inference, we resample a fixed maximum of 24 frames per video for VOS tasks. Unlike training, segmentation and evaluation are conducted over the full video sequence using SAM2 to align with standard benchmark protocols. For RefAVS, we adopt the same resampling and evaluation procedure as in VOS, ensuring consistency across task settings.

\newtcolorbox{promptbox}{
    breakable,
    enhanced,
    colback=blue!5!white,         %
    colframe=blue!40!black,       %
    coltitle=white,               %
    colbacktitle=blue!60!black,  %
    fonttitle=\bfseries\large,   %
    title=Prompt for \ours\ as System~1,
    boxrule=1.0pt,                %
    arc=6pt,                      %
    top=6pt, bottom=6pt, left=8pt, right=8pt,
    attach boxed title to top center={yshift=-1.5mm},
    before skip=10pt, after skip=10pt,
    width=\textwidth,
    boxed title style={
            colframe=blue!70!black,
            colback=blue!70!black,
            sharp corners,
            boxrule=0pt,
            top=1pt,
            bottom=1pt,
            left=4pt,
            right=4pt,
        },
    drop shadow,
}

\begin{figure}[t]
    \centering
    {\small
        \begin{promptbox}
            \begin{itemize}[leftmargin=1em, itemsep=0pt, topsep=2pt]
                \item Given a \texttt{\textcolor{red}{[frames]}} seconds video and a reference instruction: \texttt{\textcolor{red}{[ref\_prompt]}} that may involve temporal behavior, identify the exact object(s) \texttt{\textcolor{red}{[ref\_prompt]}} in the video that matches the description.
                \item Select about 4 most relevant moments that contain the referred object(s) with the best view.
                \item Then, simplify the identified object into a short and clear visual grounding description that can be used for single-image reference at each moment.
                \item Avoid temporal phrases and comparison phrases like ``walking'', ``moving'', ``approaching'', ``bigger'' or ``smaller'', but instead describe visible visual cues like clothing, pose, position, or grouping.
                \item Try to select moments that are \textbf{temporally well-distributed across the video}, rather than clustered in the same part of the timeline. Avoid selecting multiple timestamps that are adjacent or overlapping; instead, prefer clearly distinct moments that each offer unique visual information. It is better to choose the most relevant and highly representative moments \textbf{spanning the entire video}, rather than picking all from the beginning.
                \item Explain your reasoning in \textcolor{blue}{\texttt{<think></think>}} and output the final result in \textcolor{blue}{\texttt{<answer></answer>}}. Your final answer should be a JSON object in the following format:
\begin{verbatim}
<think> your analysis about the video and reference instruction </think>
<answer>
{
"start_time": "00:[start]",
"end_time": "00:[end]",
"description": "direct description of referred object(s) at this moment"
}
</answer>
\end{verbatim}
            \end{itemize}
        \end{promptbox}
    }
    \caption{Keyframe selection and recaptioning prompt for System 1.
    }
    \label{fig:sys1-prompt-vos}
\end{figure}

\begin{figure}[t]
    \centering
    {\small
        \begin{promptbox}
            \begin{itemize}[leftmargin=1em, itemsep=0pt, topsep=2pt]
                \item Given a \texttt{\textcolor{red}{[audio\_duration]}} audio and a reference instruction: \texttt{\textcolor{red}{[ref\_prompt]}}, which involves temporal and audio-related behavior, first analyze the objects in the image that are producing sound, including both human voices and instrument sounds.
                \item Based on the audio content, identify the exact object \texttt{\textcolor{red}{[ref\_prompt]}} in the image that matches the audio.
                \item Then, simplify the identified object into a short and clear visual grounding description that can be unambiguously recognized in a single image without relying on audio.
                \item Avoid using temporal expressions such as ``playing'' or ``moving''; instead, describe visible visual cues such as clothing, pose, position, or grouping.
                \item Explain your reasoning in \textcolor{blue}{\texttt{<think></think>}} and output the final result in \textcolor{blue}{\texttt{<answer></answer>}}.
            \end{itemize}
        \end{promptbox}
    }
    \caption{Audio analyzing and recaptioning prompt for System 1.
    }
    \label{fig:sys1-prompt-avs}
\end{figure}

\section{Ablation Studies}
\label{sec:appendix_b}

We conduct an ablation study to investigate the effect of progressively designed reward components $R_{\mathcal{K}}$ (keyframe coverage), $R_{\mathcal{A}}$ (alignment via Hungarian matching), and $R_{\mathcal{G}}$ (global grounding IoU) on the overall performance. To ensure a fair comparison, all models are trained for one epoch on the ReVOS and MeVIS datasets and are constrained to select exactly four keyframes unless otherwise noted.

Table~\ref{tab:reward-ablation} reports the results across four evaluation subsets of ReVOS: referring, reasoning, single-object, and multi-object. We observe that the combination of $R_{\mathcal{K}} + R_{\mathcal{G}}$ achieves the highest overall score (39.9\%), outperforming the full combination $R_{\mathcal{K}} + R_{\mathcal{A}} + R_{\mathcal{G}}$ (38.4\%). This suggests that the inclusion of $R_{\mathcal{A}}$ may introduce instability rather than improvement.

We hypothesize that this is due to the nature of $R_{\mathcal{A}}$, which relies on Hungarian matching over temporal sequences. Given that the segmentation model SAM2 already incorporates strong temporal priors, the additional alignment-based reward may not effectively capture useful gradients and could introduce variance from imperfect IoU estimation. In contrast, $R_{\mathcal{G}}$, which aggregates IoUs across frames, directly reinforces temporal consistency and spatial correctness, leading to more stable learning dynamics.

Interestingly, $R_{\mathcal{K}}$ alone provides a surprisingly strong baseline (35.2\%), demonstrating that ensuring keyframe coverage is already beneficial. However, only when combined with $R_{\mathcal{G}}$ do we observe consistent improvements across all task types, including reasoning (34.2\%) and multi-object scenarios (42.6\%).

Lastly, we include a model additionally trained with 2,000 samples from the RefCOCOg grounding dataset~\cite{liu2025segzero} as a system 2 enhanced model. This model achieves 44.7\% overall, demonstrating the potential of our design in VOS tasks lies in the grounding capabilities of the model.

These findings validate the effectiveness of $R_{\mathcal{G}}$ as a grounding-aware reward and highlight the limitations of alignment-based matching in the presence of strong perceptual priors.

\begin{table}[bt]
    \centering
    \caption{Ablation study on the reward function $R_{\mathcal{K}}$, $R_{\mathcal{A}}$ and $R_{\mathcal{G}}$ for System 1. The first model is trained with additional 2,000 samples from grounding dataset refcocog.}
    \label{tab:reward-ablation}
    \begin{tabular}{l *{6}{S}}
        \toprule
        \multirow{2}{*}{\textbf{Method}}                      & \multicolumn{5}{c}{\textbf{ReVOS}}                                                                                     \\

                                                              & \textit{Referring}                 & \textit{Reasoning} & \textit{Single} & \textit{Multi} & \textbf{\textit{Overall}} \\
        \midrule
        Omni-R1 + \textit{refcocog}                           & 52.5                               & 36.9               & 45.0            & 46.6           & 44.7                      \\
        $R_{\mathcal{K}}$+$R_{\mathcal{A}}$+$R_{\mathcal{G}}$ & 44.2                               & 32.5               & 38.2            & 41.9           & 38.4                      \\
        $R_{\mathcal{K}}$+$R_{\mathcal{G}}$                   & 45.5                               & 34.2               & 39.6            & 42.6           & 39.9                      \\
        $R_{\mathcal{K}}$+$R_{\mathcal{A}}$                   & 43.1                               & 29.5               & 36.8            & 37.5           & 36.3                      \\
        $R_{\mathcal{K}}$                                     & 44.1                               & 26.2               & 36.6            & 34.1           & 35.2                      \\

        \bottomrule
    \end{tabular}
\end{table}

\section{Visualization Results}
\label{sec:appendix_c}

\begin{figure}[bt]
    \centering
    \vspace{-1.5cm}
    \begin{subfigure}[t]{\linewidth}
        \begin{adjustwidth}{-0.9cm}{-0.9cm}
            \centering
            \includegraphics[page=1,width=\linewidth]{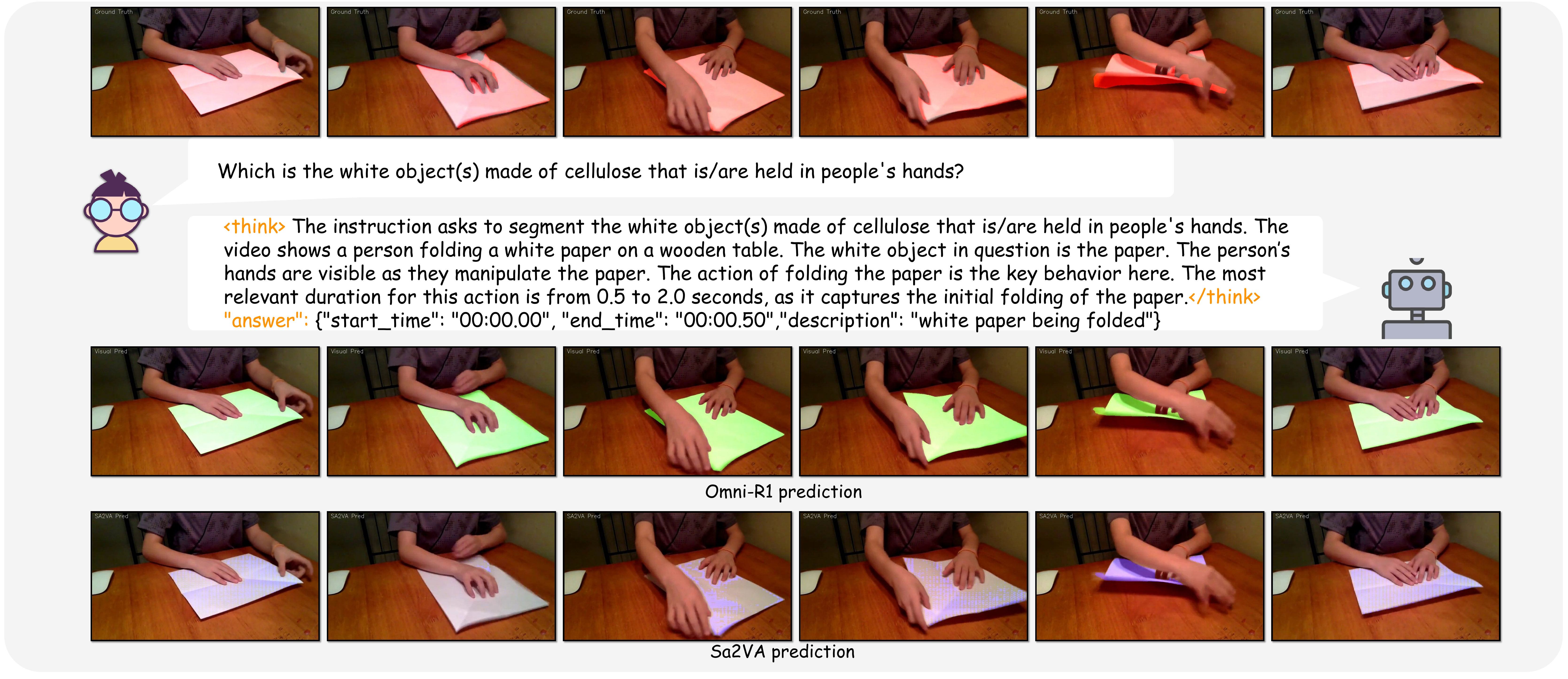}
        \end{adjustwidth}
        \vspace{-2mm}
        \caption{In this simple example, both our method and Sa2VA are able to select the right target object, but our method is able to segment the target object with a mask consistent with the ground truth, while Sa2VA generates a mask with holes and noise.}
        \label{fig:vis_1}
    \end{subfigure}

    \begin{subfigure}[t]{\linewidth}
        \begin{adjustwidth}{-0.9cm}{-0.9cm}
            \centering
            \includegraphics[page=2,width=\linewidth]{figures/vos_avs.pdf}
        \end{adjustwidth}
        \vspace{-2mm}
        \caption{The target object can only be predicted after one watches the whole video and makes a prediction according to the video context.}
        \label{fig:vis_2}
    \end{subfigure}

    \begin{subfigure}[t]{\linewidth}
        \begin{adjustwidth}{-0.9cm}{-0.9cm}
            \centering
            \includegraphics[page=3,width=\linewidth]{figures/vos_avs.pdf}
        \end{adjustwidth}
        \vspace{-2mm}
        \caption{Though our method is able to select the right wineglass at System 1, it still lets System 2 analyze the detailed information.}
        \label{fig:vis_3}
    \end{subfigure}
    \vspace{-2mm}
    \caption{Visualization results of our method on three representative VOS cases. Each subfigure illustrates a different reasoning pattern.}
    \label{fig:vis_fir}
\end{figure}

\paragraph{Mask Quality} Since our method utilizes the SAM2 model for segmentation, without fine-tuning mask decoder, the final mask output is more stable than those methods that rely on additional training on segmentation mask decoder. As can be seen in Figure \ref{fig:vis_1}, in this simple example, our method is able to segment the target object with a mask consistent with the ground truth, while Sa2VA predicts the right target but generates a mask with holes and noise.

\begin{figure}[bt]
    \centering
    \vspace{-1.5cm}
    \begin{subfigure}[t]{\linewidth}
        \begin{adjustwidth}{-1cm}{-1cm}
            \centering
            \includegraphics[page=4,width=\linewidth]{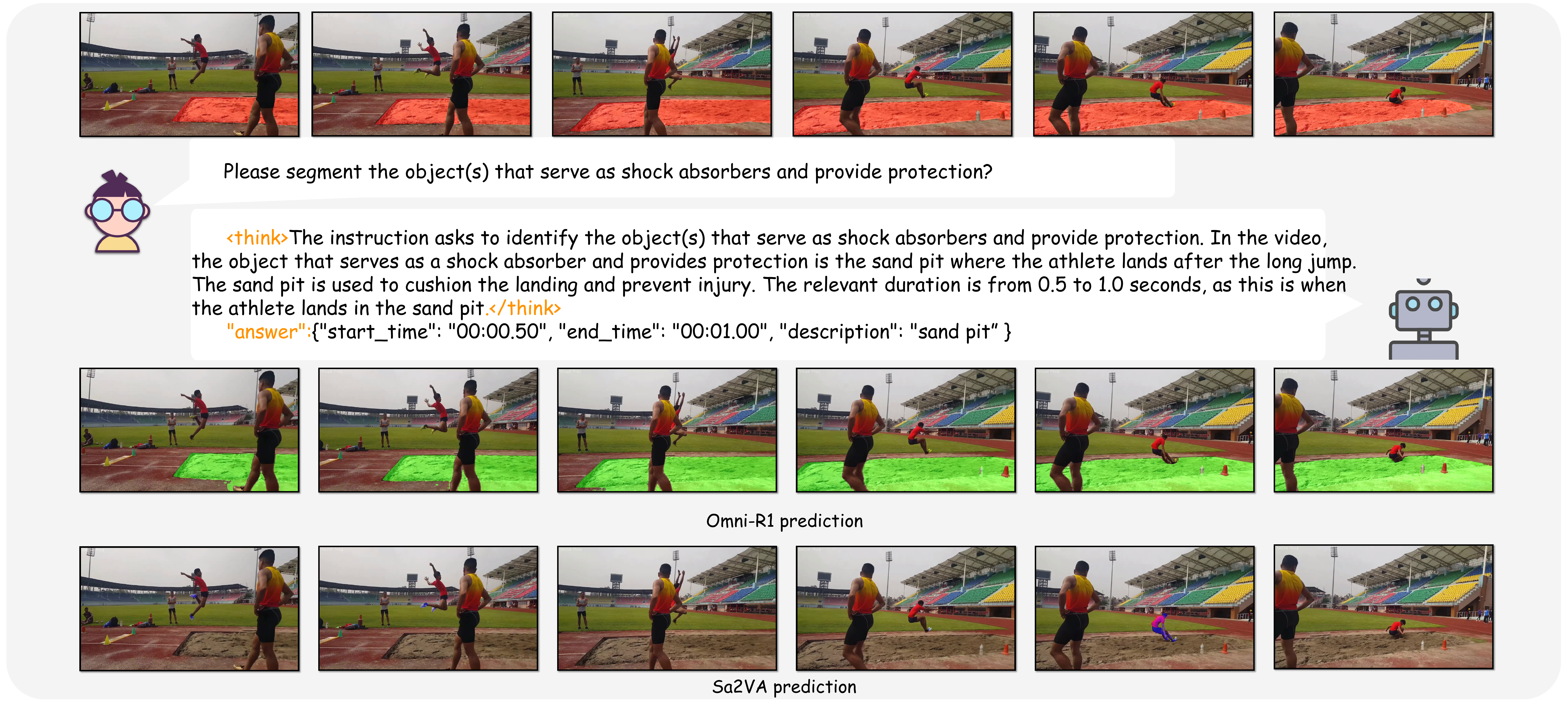}
        \end{adjustwidth}
        \vspace{-2mm}
        \caption{The case shows the video understanding abilities of our method. Our method analyses the function of the sand pit and is able to segment, while Sa2VA fails to do so.}
        \label{fig:vis_4}
    \end{subfigure}
    \begin{subfigure}[t]{\linewidth}
        \begin{adjustwidth}{-1cm}{-1cm}
            \centering
            \includegraphics[page=5,width=\linewidth]{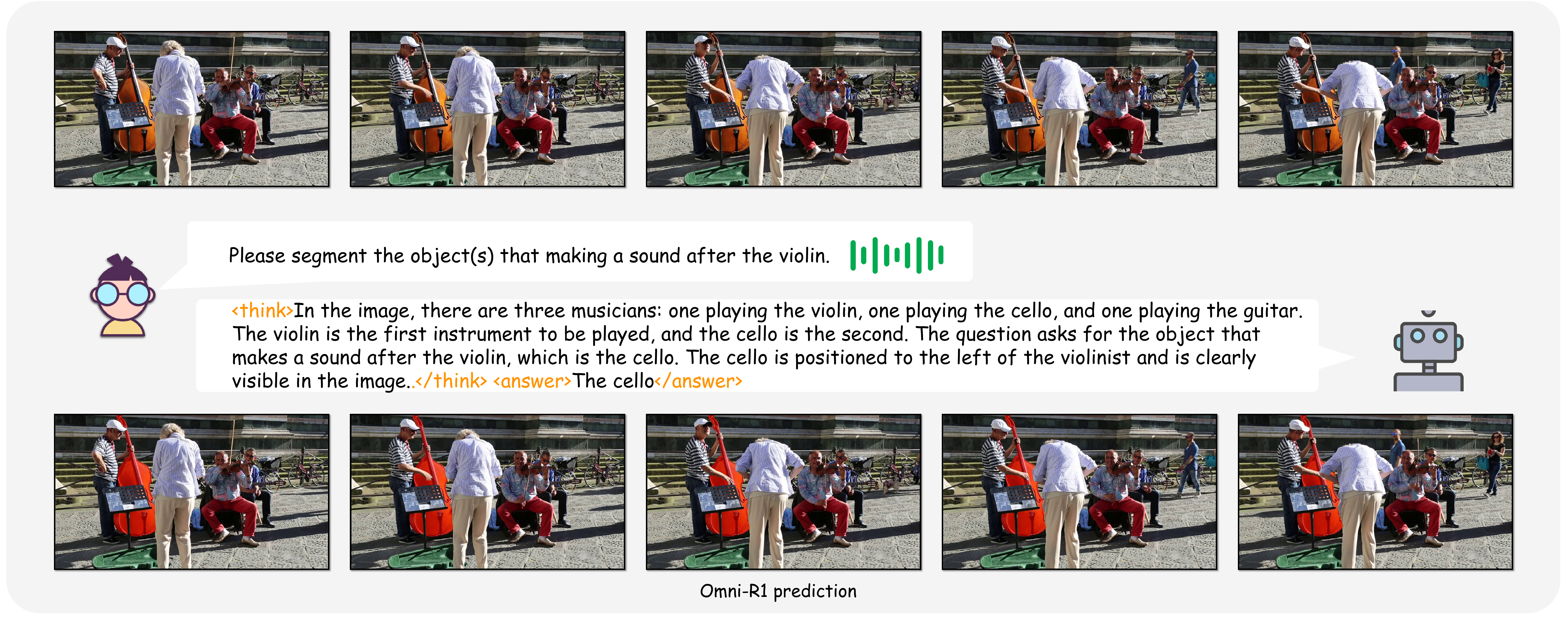}
        \end{adjustwidth}
        \vspace{-2mm}
        \caption{The case shows the video understanding abilities of our method. Our method analyses the function of the sand pit and is able to segment, while Sa2VA fails to do so.}
        \label{fig:vis_5}
    \end{subfigure}
        \begin{subfigure}[t]{\linewidth}
        \begin{adjustwidth}{-1cm}{-1cm}
            \centering
            \includegraphics[page=6,width=\linewidth]{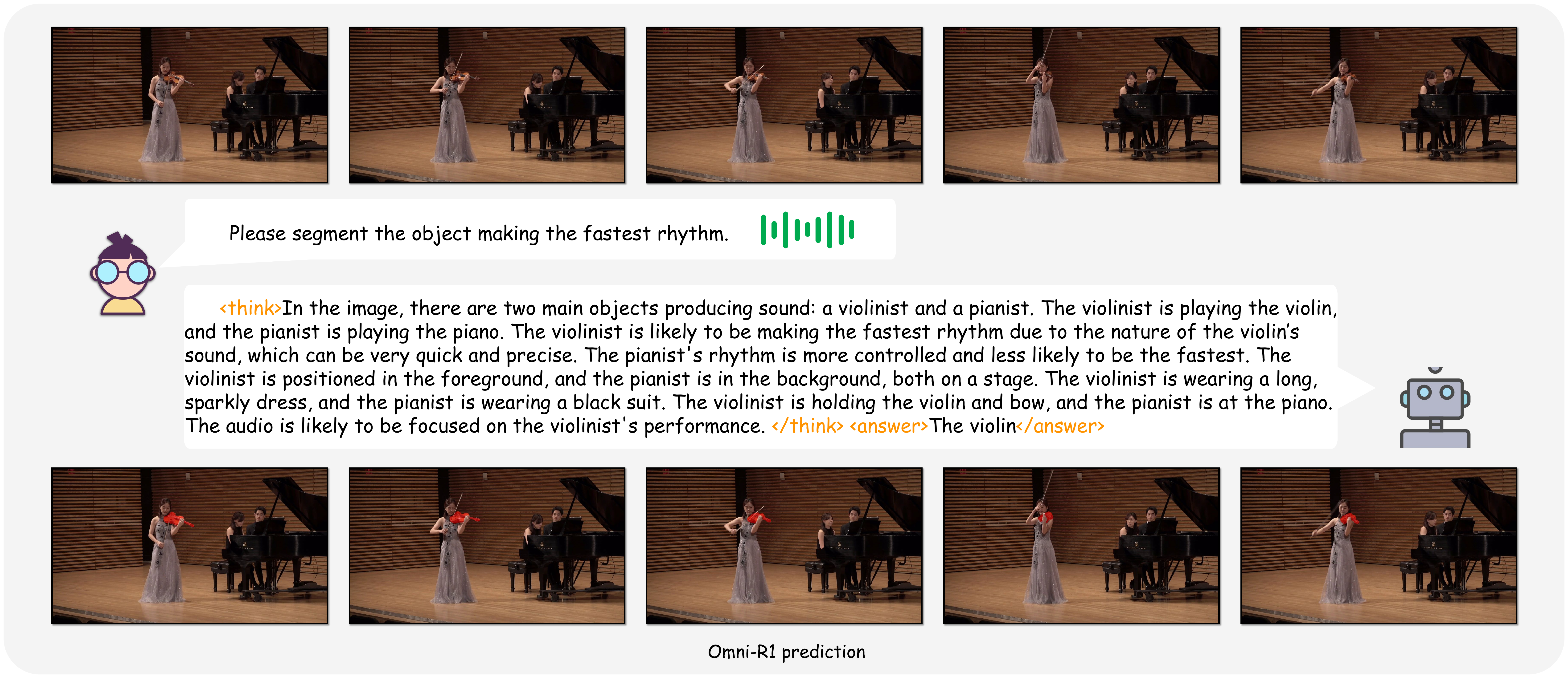}
        \end{adjustwidth}
        \vspace{-2mm}
        \caption{The case shows the video understanding abilities of our method. Our method analyses the function of the sand pit and is able to segment, while Sa2VA fails to do so.}
        \label{fig:vis_6}
    \end{subfigure}
    \vspace{-2mm}
    \caption{More visualization results of our method on representative VOS and AVS cases.}
    \label{fig:vis_sec}
\end{figure}

\paragraph{Temporal Reasoning} Our System 1 leverages temporal reasoning to improve segmentation accuracy. As can be seen in Figure \ref{fig:vis_2}, in this example, one has to \textbf{watch the whole video and analyze the video context to make a correct prediction} about the next bottle to be picked up. Our method is able to select the bottle that is about to be picked up, while Sa2VA fails to do so and segments the bottle that is already picked up. A similar case is shown in Figure \ref{fig:vis_4}, where one has to leverage world knowledge to understanding the target object and our method selects the right object while Sa2VA fails to do so. Both cases show that our method is able to leverage temporal reasoning to improve the segmentation accuracy.

\paragraph{Detail Understanding} Our System 1 leverages detail reasoning to improve segmentation accuracy. Figure \ref{fig:vis_3} shows a scenario where detail reasoning is needed to figure which wineglass is likely will be finished first. Our System 1 is \textbf{already able to select the right wineglass but still makes a loose description for System 2 to analyze detail information}, while Sa2VA fails to understand the instruction and segments all the wineglasses. This shows that our System 1 is able to delay detail reasoning for System 2 for detail understanding to improve the segmentation accuracy.

\section{More Analysis}
\label{sec:appendix_d}

\begin{table}[bt]
\centering
\caption{Video Object Segmentation performance on MeVIS across different methods, the metric is \njf\ score(\%). $\ddag$ means the results are evaluated where Omni-R1-7B serves as System \num{1} and Sa2VA-1B as System \num{2}}
\label{tab:mevis-comparison}
\begin{tabular}{l *{2}{S}}
\toprule
\multirow{2}{*}{\textbf{Model}} &  \textbf{MeVIS} \\

& \textbf{\textit{val\_u}} \\
\midrule

Sa2VA-1B~\cite{yuan2025sa2va} &  \num{53.4} \\
Sa2VA-4B~\cite{yuan2025sa2va} &  \num{55.4} \\
\midrule
$\text{Qwen2.5-Omni-7B}^{\dag}$ & \num{33.6} \\
% $\text{Omni-R1-7B}^{\dag}$ & \num{34.9} \\
$\text{Omni-R1-8B}^{\ddag}$ & \bfseries \num{55.4} \\
% $\text{Omni-R1-11B}^{\ddag}$ & \num{54.6} \\
\bottomrule
\end{tabular}
\end{table}

\paragraph{Hallucination Analysis}

During the training of our RefAVS task, we identified a significant hallucination problem, which we attribute to the complexity of multi-modal video and audio inputs. To systematically evaluate this issue, we conducted targeted assessments on audio-related hallucinations using the JUDGE subset of AVHBench~\cite{sung2024avhbench}, the first comprehensive benchmark designed to evaluate the perception and comprehension abilities of audio-visual large language models (LLMs).

As shown in Table~\ref{tab:avhbench-results}, our base model (Qwen2.5-Omni-7B) achieves an accuracy of 58.5\% on the JUDGE subset. Training on 1600 AVS samples leads to a modest improvement (60.8\%), which is further enhanced to 61.5\% by applying the GRPO KL loss with a reduced coefficient ($\beta = 0.001$). Notably, increasing the AVS training samples to 10400 does not yield better results, suggesting potential overfitting or task imbalance.

On the other hand, training with VOS tasks alone significantly boosts accuracy to 66.0\%, and the best performance (71.9\%) is obtained by jointly training on both AVS and VOS tasks. This represents a substantial improvement of 13.4\% over the base model, demonstrating that multi-task training not only enhances audio-visual grounding but also mitigates hallucination issues more effectively.

These results confirm the effectiveness of leveraging task diversity and balanced reward optimization in improving the robustness of multimodal reasoning.

\begin{table}[bt]
    \centering
    \caption{Performance on AVHBench (JUDGE subset, total 5302 samples). In the table, AVS tasks are trained on RefAVS dataset and VOS tasks are trained on ReVOS and MeViS datasets. The default GRPO KL loss weight $\beta = 0.04$.}
    \label{tab:avhbench-results}
    \begin{tabular}{lcc}
        \toprule
        \multirow{2}{*}{\textbf{Method}}      & \multicolumn{2}{c}{\textbf{AVHBench JUDGE}}            \\
                                              & Correct Answers                             & Accuracy \\
        \midrule
        Base Model                            & 3100                                        & 58.5\%  \\
        AVS 1600 samples                      & 3222                                        & 60.8\%  \\
        AVS 1600 samples with $\beta = 0.001$ & 3261                                        & 61.5\%  \\
        AVS 10400 samples                     & 3120                                        & 58.9\%  \\
        VOS                                   & 3500                                        & 66.0\%  \\
        AVS and VOS                           & 3811                                        & \bfseries 71.9\%  \\
        \bottomrule
    \end{tabular}
\end{table}

\paragraph{Video Resolution Influence on General Video Understanding Tasks}

\begin{table}[bt]
    \centering
    \caption{Performance comparison across different resolutions and the use of a thinking prompt on VideoMME and MVBench. Resolutions are set to either \texttt{128$\times$28$\times$28} (default) or \texttt{256$\times$28$\times$28} (high). The thinking prompt provides an additional reasoning cue. The reported metric is the average of $\mathcal{J}$ and $\mathcal{F}$ scores (\%).}

    \label{tab:resolution}
    \begin{tabular}{lccccc}
        \toprule
        \multirow{2}{*}{\textbf{Model}} & \multirow{2}{*}{\textbf{Resolution}} & \multirow{2}{*}{\textbf{Thinking}} & \multicolumn{2}{c}{\textbf{VideoMME}} & \textbf{MVBench}                \\
                                        &                                      &                                    & \textbf{General}                      & \textbf{Short}   & \textbf{Avg} \\
        \midrule
        Qwen2.5-Omni                    & 128×28×28                            & No                                 & 58.3                                  & 69.8             & 66.1         \\
        Qwen2.5-Omni                    & 256×28×28                            & No                                 & 58.7                                  & 69.9             & 67.0         \\
        Qwen2.5-Omni                    & 128×28×28                            & Yes                                & 59.3                                  & 70.1             & 68.1         \\
        Qwen2.5-Omni                    & 256×28×28                            & Yes                                & 59.8                                  & 70.9             & 68.3         \\
        \midrule
        Omni-R1-AVS                     & 128×28×28                            & No                                 & 59.0                                  & 71.9             & 68.3         \\
        Omni-R1-AVS                     & 256×28×28                            & No                                 & 59.4                                  & 71.9             & 68.7         \\
        Omni-R1-AVS                     & 128×28×28                            & Yes                                & 59.9                                  & 72.1             & 69.4         \\
        Omni-R1-AVS                     & 256×28×28                            & Yes                                & 60.0                                  & 72.1             & 69.5         \\
        \midrule
        Omni-R1-VOS                     & 128×28×28                            & No                                 & 59.7                                  & 72.3             & 68.9         \\
        Omni-R1-VOS                     & 256×28×28                            & No                                 & 59.6                                  & 72.5             & 68.9         \\
        Omni-R1-VOS                     & 128×28×28                            & Yes                                & 59.8                                  & 72.5             & 69.8         \\
        Omni-R1-VOS                     & 256×28×28                            & Yes                                & 60.1                                  & 72.8             & 69.9         \\
        \midrule
        Omni-R1-VOS-AVS                 & 128×28×28                            & No                                 & 60.1                                  & 72.5             & 69.1         \\
        Omni-R1-VOS-AVS                 & 128×28×28                            & Yes                                & 60.7                                  & 73.0             & 70.3         \\

        \bottomrule
    \end{tabular}
\end{table}

To evaluate the influence of input resolution and prompting strategy on general video understanding, we compare model performance across different configurations on the VideoMME and MVBench benchmarks, as summarized in Table~\ref{tab:resolution}. All models are evaluated under two resolution settings: the default resolution of \texttt{128$\times$28$\times$28} and a higher resolution of \texttt{256$\times$28$\times$28} (denoted with *), with and without the proposed \textit{thinking} prompting strategy.

We observe that increasing the input resolution consistently leads to performance gains across all models. For instance, Qwen2.5-Omni improves from 66.1\% to 67.0\% on MVBench when evaluated at higher resolution. Similarly, our Omni-R1-AVS model benefits from the resolution increase, achieving a performance gain from 68.3\% to 68.7\%. These improvements suggest that higher spatial resolution enhances the model's ability to capture fine-grained visual details, particularly beneficial for multi-object reasoning and scene comprehension.

In addition to resolution, the \textit{thinking} prompt designed to guide the model toward structured multi-step reasoning further boosts performance across all tested models. Omni-R1-AVS with \textit{thinking} achieves 69.4\% on MVBench, outperforming its baseline by 1.1\%. The combination of both higher resolution and \textit{thinking} yields the best results overall, with Omni-R1-VOS-AVS + \textit{thinking}* reaching 60.7\% on VideoMME (general) and 70.3\% on MVBench. This indicates that resolution and prompting act as complementary strategies: resolution improves visual precision, while prompting enhances reasoning capability.

However, the performance gain obtained by increasing video resolution is marginal, suggesting that in general understanding benchmarks, resolution plays a limited role, and temporal understanding is more critical than fine-grained spatial details. This observation is consistent with our findings in the main paper, where the dual-system design significantly enhances the model’s temporal reasoning capabilities and yields notable improvements on reasoning-intensive VOS tasks.

\section{Limitations and Future Work}
\label{sec:appendix_e}

\paragraph{Limitations}
Although our dual-system design significantly enhances the temporal reasoning capability of \textbf{System 1}, the complete functional decoupling between System 1 and System 2 introduces certain limitations. In particular, \textbf{System 2} lacks temporal context, which may affect consistency in temporally coherent tasks. This consideration partially motivates our selection of \textbf{VOS} as a primary training task: while VOS emphasizes temporal consistency, it also provides dense per-frame annotations that allow us to design stable training strategies to mitigate the context gap—such as frame-wise Hungarian matching loss and aggregated mask-based rewards. During inference, the missing temporal cues in System 2 are partially recovered through SAM2's mask-based processing.

However, when extending to tasks requiring finer temporal sensitivity such as detecting and describing localized anomalous behaviors within a specific time span, our current architecture faces new challenges. While \textbf{System 1} can still progressively narrow down relevant temporal segments, the need for dynamic temporal granularity exposes the limitations of a fully decoupled, one-way reasoning architecture.

\paragraph{Discussion on Future Work}
Our coarse-to-fine reasoning pathway aligns closely with human cognitive intuition, yet we recognize the crucial importance of enabling a bidirectional flow of information between global and local levels. Our current design can be viewed as a context-constrained tree search structure, where only the root node (\textbf{System 1}) has access to full contextual information, while downstream nodes (\textbf{System 2}) operate solely on partial, local input. This diverges from typical hierarchical systems, which often allow child nodes to access aggregated information from their parent nodes.

While the VOS task structure inherently compensates for this limitation by providing dense temporal supervision, to further enhance the flexibility of our two-system framework, future work should explore more interactive architectures that facilitate explicit information exchange between \textbf{System 1} and \textbf{System 2}. Supporting backtracking within the reasoning tree would empower \textbf{System 1} to refine its global reasoning using local insights from \textbf{System 2}. Conversely, it would enable \textbf{System 2} to operate with broader contextual awareness provided by \textbf{System 1}. We firmly believe that transforming this pipeline into a bidirectional, cooperative reasoning structure holds immense potential for advancing multi-modal temporal understanding abilities on more flexible and complex tasks.

\clearpage
\medskip
\bibliographystyle{unsrt}       %
\bibliography{refs}
\clearpage

\end{document}